\documentclass[subscriptcorrection,upint,varvw,balance,hyphenate,nofoot,nolists]{asmejour}

\hypersetup{
	pdfauthor={Qianwen Zhao, Long Wang},
	pdftitle={Predicting Grasping Compliance in Robotic Hands through Analytical-Model-Informed Neural Networks},
	pdfkeywords={tool grasping stability, grasping compliance, analytical-model-informed learning, underactuated robotic hand},
	pdfsubject={Manuscript preprint}
}

\begin{document}

\SetAuthorBlock{Qianwen Zhao}{Department of Mechanical Engineering,\\
Stevens Institute of Technology,\\
Hoboken, NJ, USA\\
email: qzhao10@stevens.edu}

\SetAuthorBlock{Long Wang\CorrespondingAuthor}{Department of Mechanical Engineering,\\
Stevens Institute of Technology,\\
Hoboken, NJ, USA\\
email: lwang4@stevens.edu}

\title{Predicting Grasping Compliance in Robotic Hands through Analytical-Model-Informed Neural Networks}
\begin{abstract}
In robotic manipulation studies, grasping is often treated as a binary success or failure problem, usually defined by whether the object simply stays in the hand.
For forceful tool use, however, this view is insufficient because grasp compliance becomes a critical factor governing how the hand and tool behave under load.
Compliance arises from coupled kinematics, grasp configuration, passive mechanics, and contact conditions, producing nonlinear behavior in which deformation and interaction forces influence each other.
Understanding this relationship is essential for predictive models of how a grasped tool and a compliant hand jointly respond to external loading.
In underactuated hands, these effects are amplified: such designs offer low cost and adaptive grasping, but make compliance behavior more difficult to model and predict.
Our goal is therefore to develop a predictive model for grasped tool behavior during forceful interactions.
To address this challenge, we introduce an analytical model informed neural network (AMINN), a hybrid predictive model that combines an analytical mechanics layer with data driven learning to estimate grasp stability and in hand tool displacement under external loading.
The model is evaluated on a three finger underactuated robotic hand and shows strong predictive capability with mechanically meaningful outputs across diverse loading conditions.
Compared with a black box multilayer perceptron baseline, AMINN also achieves better energy based physical consistency.
Beyond prediction accuracy alone, this framework advances physically interpretable learning for robotic manipulation and supports more reliable, safer, and more trustworthy autonomous tool use in safety critical settings during forceful interaction.
\end{abstract}

\date{}
\maketitle

\section{Introduction} \label{sec:introduction}

Robotic manipulation is increasingly deployed in settings that require reliable interaction with tools and objects, including industrial assembly, maintenance operations, and medical procedures. In these applications, the hand is not only responsible for grasp retention, but also for transmitting force in task-relevant directions while preserving positional accuracy of the grasped tool. As robotic systems move from transport-focused manipulation toward forceful, contact-rich interaction, grasp behavior becomes a central determinant of both safety and performance.

Most grasp-evaluation pipelines still emphasize binary outcomes such as stable/unstable or success/failure. This perspective is useful in pick-and-place tasks, where the primary objective is preventing object drop. However, for tool use, binary retention alone is often insufficient. A grasp may remain nominally stable while still allowing in-hand tool displacement large enough to degrade trajectory tracking, reduce task precision, or compromise safety margins. In these scenarios, the practical question is not only whether the tool is retained, but also how it deforms and moves under applied wrench.

\begin{figure}[!t]
	\centering
	\includegraphics[width=0.99\columnwidth]{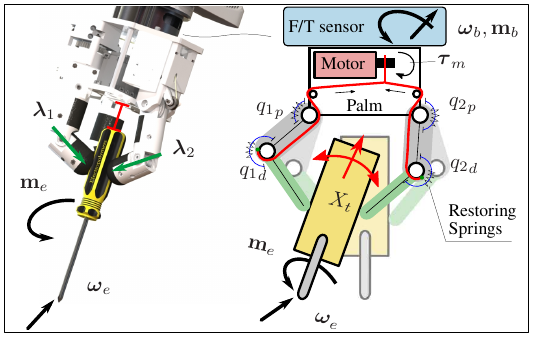}
	\caption{Underactuated robotic hand grasping a screwdriver-like tool before and after loading, illustrating grasping compliance during tool use.}
	\label{fig:grasping-compliance-pinch-before-after}
\end{figure}

Figure~\ref{fig:grasping-compliance-pinch-before-after} illustrates this motivation in a representative underactuated hand grasping a screwdriver-like tool: under loading, the tool exhibits measurable in-hand displacement even when grasp retention is preserved. This behavior is especially important in underactuated mechanisms, where passive elasticity, tendon routing, and finger coupling interact with contact conditions in a configuration-dependent way. Consequently, the wrench-to-displacement mapping is nonlinear, state dependent, and strongly influenced by grasp pose and load direction.

Existing compliance-modeling strategies each provide part of the solution, but leave important gaps for practical forceful tool manipulation. Analytical models offer physical interpretability and explicit structure, yet often depend on local linearization, idealized contact assumptions, and difficult-to-identify stiffness parameters; moreover, as robotic systems become more complex, deriving and maintaining a complete analytical model is often difficult or not always practically available. High-fidelity simulation and finite-element pipelines can capture richer deformation behavior, but usually require substantial calibration effort and computational resources. Purely black-box learning can model nonlinear mappings directly, but this flexibility often comes with weak physical interpretability and higher risk of non-physical hallucinated predictions under changing grasps, tools, and loading directions. In forceful contact-rich tasks, these failure modes are not only modeling artifacts: they can degrade execution quality, reduce operator trust, and create real safety risk for both the robot and nearby humans when predicted behavior is physically implausible.

To address this gap, we propose an \emph{Analytical Model-Informed Neural Network (AMINN)}, a hybrid method that combines analytical modeling and learning for underactuated robotic hand tool use. AMINN integrates a structured analytical mechanics layer into a two-stage cascade prediction pipeline: Stage 1 predicts whether the grasp remains stable under external loading, and Stage 2 predicts in-hand tool displacement only for samples that pass stability prediction. Instead of treating load-response prediction as unconstrained black-box regression, AMINN enforces mechanically grounded structure inside the predictor while still learning state-dependent nonlinear behavior from data.

This design targets a practical objective for forceful manipulation: preserve strong predictive capability while producing physically meaningful and physically consistent predictions that are more trustworthy for deployment in safety-critical robotic operation. The main contributions of this work are:
\begin{itemize}
    \item[1.] We propose an analytical model-informed neural network (AMINN) to predict the tool-handling behavior of underactuated robotic hands under external loading.
    \item[2.] We integrate a structured analytical mechanics layer into AMINN, enabling mechanically meaningful predictions while promoting physical consistency.
    \item[3.] We demonstrate that AMINN preserves strong predictive capability while achieving energy-based physical consistency relative to a plain multilayer perceptron baseline.
\end{itemize}

\section{Related Work}

\subsection{Grasp Stability and Binary Success Metrics}
Grasp stability has long been a central objective in robotic manipulation. Foundational treatments of hand mechanics and multifingered grasping formalized contact constraints, internal forces, and object motion inside articulated hands \cite{MasonSalisbury1985, KerrRoth1986, BrodskyShoham1995}. Classical analytical approaches then evaluate whether a grasp can resist disturbances using force-closure and wrench-space criteria \cite{Nguyen1986, Ferrari1992}, often together with quality measures that rank robustness margins \cite{Roa2014}. These formulations remain foundational in grasp analysis and synthesis because they make explicit whether a contact configuration can resist disturbance wrenches at all.

In parallel, many practical systems adopt empirical binary labels, where a grasp is considered successful if the object can be lifted and retained without slip or drop. This framing enabled the rise of large-scale learning-based grasp prediction from vision and interaction data, including deep grasp detection from RGB-D input \cite{Lenz2015RoboticGrasps}, large-scale self-supervised grasp success learning \cite{Pinto2016, levine2018learning}, surface-patch or grasp-geometry robustness prediction \cite{Seita2016SIMPAR_SurfacePatchPairs}, synthetic-data grasp ranking pipelines such as Dex-Net 2.0 \cite{Mahler2017DexNet2}, closed-loop reinforcement-learning approaches such as QT-Opt \cite{Kalashnikov2018QTOpt}, and tactile stability classification \cite{Mi2021RCAR_TactileGCNStability}. These methods are highly effective when the main question is whether a candidate grasp will succeed or fail under a pickup-style disturbance.

For forceful tool manipulation, however, binary stability alone is often too coarse. A grasp can remain stable while still allowing in-hand displacement that is large enough to degrade precision and task success. Binary grasp-success prediction is therefore effective for transport and screening, but it does not model wrench-conditioned in-hand displacement during tool use.

\subsection{Stiffness Analysis of Robotic Mechanisms}
Stiffness analysis is typically formulated as a local Cartesian mapping between differential displacement and applied wrench \cite{StiffnessAnalysisMethodsSummary2016, Gosselin1990TRO}. Related studies on parallel robots also examined Jacobian conditioning, manipulability, and accuracy as design-relevant performance indicators \cite{Merlet2006JMD}, as well as design tradeoffs in new six-degree-of-freedom parallel architectures \cite{SimaanGlozmanShoham1998}. Widely used modeling families include Virtual Joint Method (VJM) or lumped compliance models \cite{Tsai1999RobotAnalysis}, Matrix Structural Analysis (MSA) \cite{MSA2019}, and Finite Element Analysis (FEA) \cite{Bouzgarrou2004}. VJM provides compact parameterization, MSA adds structural detail through assembled element models, and FEA provides higher geometric and material fidelity at increased computational and modeling cost. Related formulations also accounted explicitly for elastic structural deformation in parallel mechanisms \cite{Yoon2002}.

Beyond method taxonomies, several studies clarified how stiffness should be decomposed into physically distinct contributions. Cutkosky and Kao separated robotic hand compliance into servo-driven and structural components \cite{Kao1989TRA}. Chen and Kao developed conservative congruence transformations between joint and Cartesian stiffness matrices for robotic hands and fingers \cite{HandObjectStiffness2000}. Pashkevich \emph{et al.} showed how passive joints reshape manipulator stiffness modeling and projection \cite{Pashkevich2011}. Reviews of redundancy in parallel mechanisms further summarized how kinematic and actuation redundancy affect singularity avoidance, force transmission, and design options \cite{SchreiberGosselin2018}, while dedicated treatments of redundant actuation highlighted internal-force redistribution as an additional design lever \cite{Mueller2008RedundantActuation}. In parallel robots, Simaan and Shoham formulated stiffness synthesis for variable-geometry mechanisms \cite{SimaanShoham2003}, while later work examined configuration-dependent stiffness controllability in cable-driven systems \cite{Cui2021}, the effects of cable elasticity and mass on static and dynamic stiffness \cite{Yuan2015}, and directional stiffness modulation with kinematic redundancy and variable stiffness joints \cite{OrekhovSimaan2019}.

These studies provide the analytical vocabulary of Cartesian stiffness, active/passive or structural/controlled contributions, and directional modulation. However, direct transfer to underactuated hand--tool interaction remains challenging. Effective stiffness varies with configuration, load direction, and contact regime, and complete analytical modeling may become unavailable or intractable as system complexity grows. This motivates structured analytical model-informed learning that preserves mechanics while adapting to data. 

\subsection{Underactuated Hand Mechanics Modeling}
The behavior of underactuated hands is tightly coupled to tendon routing, differential transmission, springs, passive joints, and contact sequencing, so grasp mechanics cannot be separated from hand design itself \cite{LaliberteGosselin1998, Dollar2006IDETC, Gosselin2002, UnderactuatedFingers2007, CompliantHandsSurvey2011}. This literature established that underactuation trades full independent control for adaptive contact formation, lower actuation count, and simpler packaging, but also makes load sharing and post-contact behavior strongly configuration dependent.

Several works studied this mechanics more directly through quasistatic and grasp-analysis tools. Ciocarlie and Allen proposed a design and analysis framework for underactuated compliant hands that predicts slip, unbalanced object forces, and grasp stability during grasp execution \cite{CiocarlieAllen2009}. Kragten and Herder introduced performance metrics for the ability of underactuated hands to grasp and hold disturbed objects \cite{KragtenHerder2010}. Odhner and Dollar further showed that underactuated hands can support stable open-loop precision manipulation despite limited actuation \cite{UnderactuatedStiffness2015}.

Compliance-oriented studies brought this discussion closer to the present problem. Malvezzi and Prattichizzo evaluated grasp stiffness in underactuated compliant hands \cite{Domenico2013ICRA}, and Chen \emph{et al.} optimized tendon-driven hand underactuation through mechanically realizable manifolds in posture and torque spaces so that grasp postures and quasistatic loading behavior are co-shaped by design \cite{Chen2020TRO}. At the sensing level, underactuated hands have also been paired with joint-angle and force sensing to recover useful state feedback without the cost and packaging burden of densely instrumented fully actuated hands \cite{Wang2011HandSensors}. Taken together, these studies explain important pieces of underactuated-hand mechanics and design, but they stop short of a hybrid predictor that embeds analytical stiffness structure inside a learned model for wrench-conditioned tool displacement and stability.

\subsection{Learning-Based Compliance Estimation}
Learning-based methods are increasingly used to infer compliance-related quantities from interaction signals, including object property inference from grasp interaction \cite{Kutsuzawa2024}, data-driven transition modeling for compliant underactuated manipulation \cite{Morgan2019}, and physics-guided deformation prediction in compliant robotic systems \cite{Wang2024PINNRay}. Related optimization-based work in dexterous hand manipulation also emphasized how difficult it is to reason through rapidly changing contact sequences, rolling, and sliding interactions even without explicit compliance estimation \cite{Mordatch2012ContactInvariant}. These approaches demonstrate that interaction-rich hand behavior is difficult to model analytically and that learning or optimization can capture nonlinear effects that are otherwise hard to specify from first principles.

Purely black-box learning remains attractive for modeling flexibility, but it often offers limited mechanical interpretability and can produce non-physical hallucinated outputs, which reduces trust in safety-critical manipulation. At the same time, physics-guided and hybrid learning methods provide an important direction by embedding physical structure to improve plausibility and robustness. Building on this direction, our AMINN framework emphasizes joint prediction of stability and in-hand displacement, while explicitly evaluating physical consistency/passivity alongside predictive performance.

\section{Problem Statement} \label{sec:problem-statement}

\begin{figure}[!t]
	\centering
	\includegraphics[width=0.85\columnwidth]{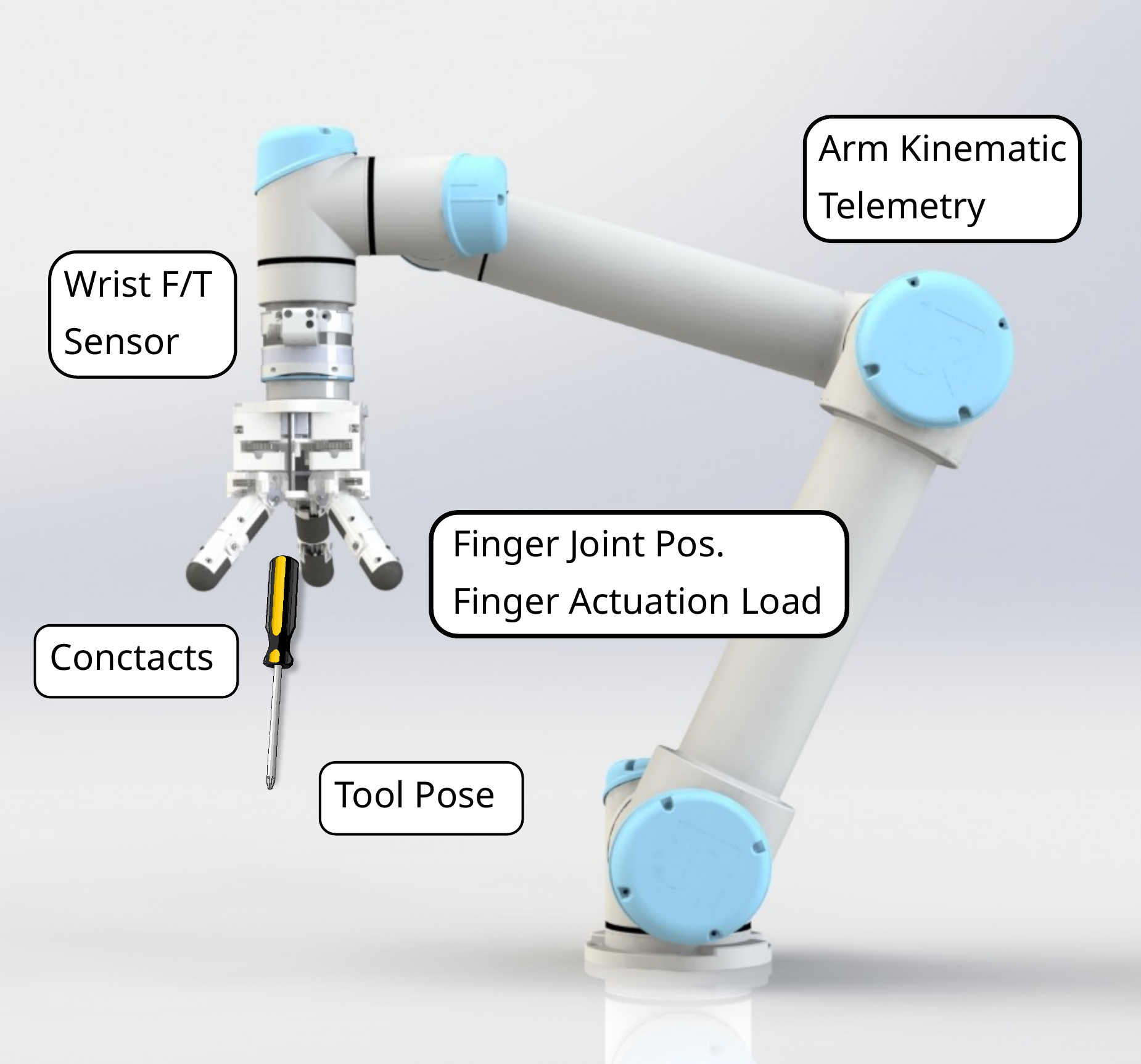}
	\caption{Representative arm--wrist--hand stack highlighting commonly available sensing cues used to estimate the state: robot-arm kinematics, wrist force/torque sensing, and sparse hand-side signals such as finger position and actuator- or tendon-side load information.}
	\label{fig:sensing-stack}
\end{figure}

\begin{figure*}[!ht]
	\centering
	\includegraphics[width=1.99\columnwidth]{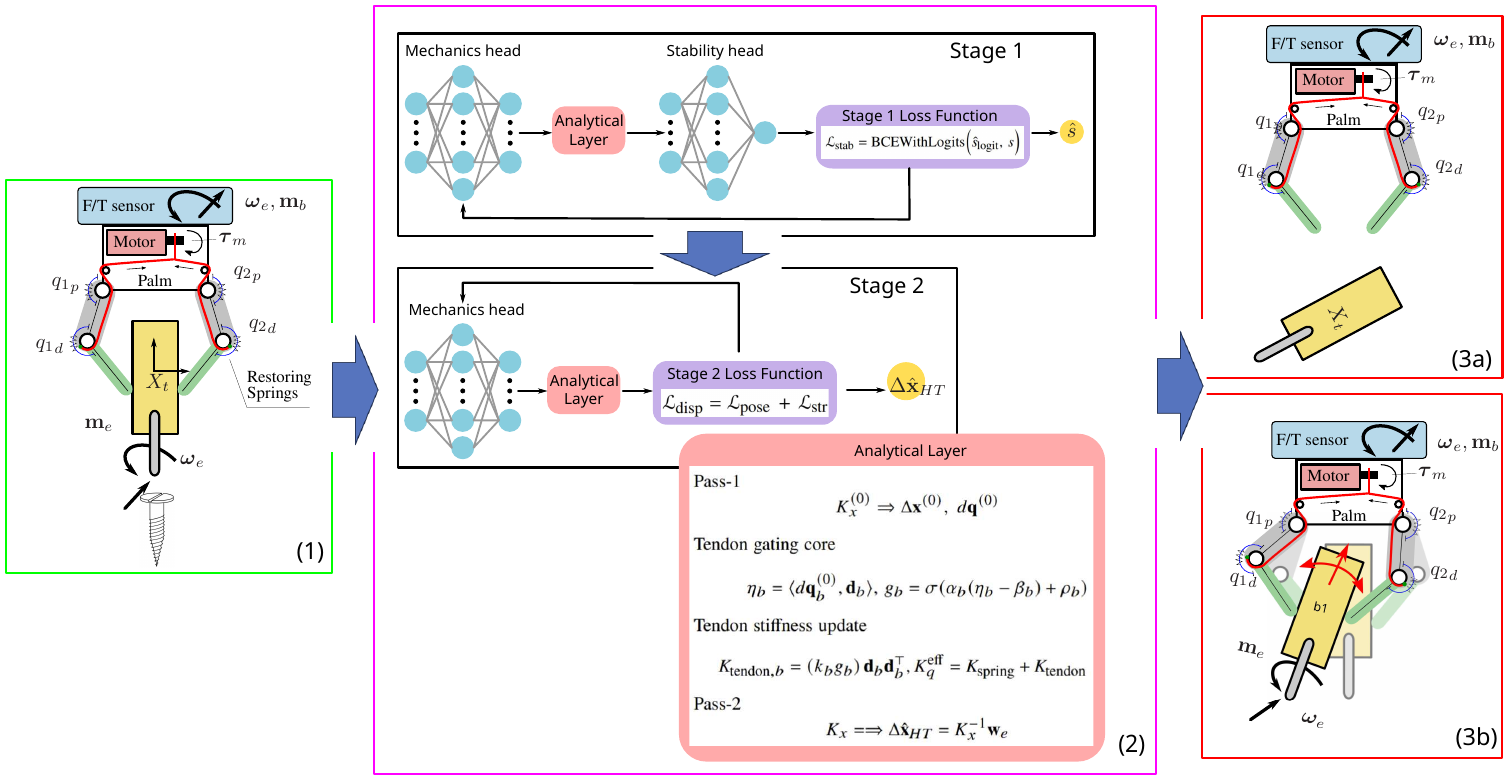}
	\caption{Overview of the prediction pipeline.
(1) A robotic hand grasps a tool, and the grasped tool undergoes loading;
(2) the proposed analytical-model-informed learning architecture;
(3) model outputs: (a) binary stability and (b) tool displacement.}
	\label{fig:problem-statement}
\end{figure*}

To formalize the objective introduced in Sec.~\ref{sec:introduction}, we consider forceful tool-use settings where a robotic hand mounted on a robot arm first grasps a tool and then uses that tool to interact with the environment. The external load applied at the tool is transmitted back through the grasp and into the hand-arm system. As a result, binary retention alone is not sufficient for task success: a grasp can remain nominally stable while still allowing enough in-hand tool motion to degrade force transmission, trajectory accuracy, or safety margin.

This problem is also practically motivated by sensing availability. Figure~\ref{fig:sensing-stack} highlights a representative arm--wrist--hand stack and the sensing cues typically available in practice. Wrist force/torque sensing at the robot arm is now common and directly supports estimation of the external wrench acting on the tool. In cable-driven hands, tendon tension or actuator-side load information is often available as an additional cue about transmission state. Fully actuated hands with dense sensing can provide richer state feedback, but they are often expensive, bulky, and difficult to package. Underactuated hands usually rely on sparser instrumentation, so low-cost joint or finger position sensing becomes especially valuable. Taken together with contact-state estimation, these signals motivate a state representation that mixes estimated quantities from sensing and geometry with analytical quantities from the mechanics model.

We define the observable state of one sample as
\begin{equation}
\mathbf{z}=\left(\mathbf{q},\ \mathbf{x}_{HT},\ \mathbf{w}_e,\ \mathcal{C},\ J_x,\ J_q\right),
\end{equation}
where $\mathbf{q}$ is hand joint configuration, $\mathbf{x}_{HT}\in\mathbb{R}^6$ is nominal hand-to-tool pose, $\mathbf{w}_e\in\mathbb{R}^6$ is external wrench, $\mathcal{C}$ is an estimated contact set, and $J_x,J_q$ are closure Jacobian terms used by the mechanics-informed predictor. The first four variables are quantities obtained from sensing, kinematic estimation, and contact inference, and the latter two variables are Jacobian terms derived from the analytical-hand-tool model. 

The prediction target is defined in a cascade form. Stage 1 estimates grasp stability,
\begin{equation}
\hat{s}=\mathcal{F}_{\mathrm{stab}}(\mathbf{z}),\qquad \hat{s}\in[0,1],
\end{equation}
and Stage 2 predicts in-hand displacement
\begin{equation}
\Delta \hat{\mathbf{x}}_{HT}=\mathcal{F}_{\mathrm{disp}}(\mathbf{z}),\qquad \Delta \hat{\mathbf{x}}_{HT}\in\mathbb{R}^6,
\end{equation}
which is used for samples that pass the stability decision threshold. This formulation directly matches the deployment objective: first determine whether the grasp remains functional, then estimate displacement for stable cases.

We work in a quasi-static regime with rigid bodies and Coulomb-frictional contacts; high-speed impact transients are out of scope. Under these assumptions, the dominant challenge is configuration- and load-dependent compliance in underactuated grasping, where passive joint elasticity, tendon effects, and contact conditions interact to produce strongly nonlinear behavior.

Figure~\ref{fig:problem-statement} summarizes this problem pipeline from grasp state and loading to cascade outputs (stability and in-hand displacement). This problem definition motivates a predictor that is both accurate and mechanically grounded, so outputs remain physically plausible and useful for safety-critical robotic decision making.

\section{Analytical Model-informed Learning Approach}
\label{sec:nn}

\subsection{Modeling of Compliance}

In the forceful manipulation setting of Sec.~\ref{sec:problem-statement}, the key quantity is the local mapping from external wrench variation to in-hand tool displacement. Under quasi-static conditions with rigid bodies and Coulomb friction, this mapping is represented by the Cartesian stiffness matrix:
\begin{equation}
    K_x = \frac{\partial \mathbf{w}}{\partial \mathbf{x}},
    \label{eq:general-stiffness}
\end{equation}
where $\mathbf{w}\in\mathbb{R}^6$ is wrench and $\mathbf{x}\in\mathbb{R}^6$ is hand-relative tool pose. Locally, $\Delta \mathbf{w}\approx K_x \Delta \mathbf{x}$.

We model the grasped hand--tool system as a parallel mechanism because fingers are mechanically coupled through a common grasped tool under contact constraints. In this viewpoint, the tool behaves as the moving platform and the palm behaves as the fixed base, while each finger acts as one kinematic branch. This formulation is useful for forceful in-hand loading because it directly captures cross-finger coupling and the configuration dependence of grasp compliance in a structured way.

Let $\mathbf{q}$ denote joint coordinates. The linearized closure and statics relations are written as
\begin{equation}
    J_x \, d\mathbf{x} + J_q \, d\mathbf{q} = 0,
\end{equation}
\begin{equation}
    d\mathbf{w} = J^\top d\boldsymbol{\tau}, \qquad d\boldsymbol{\tau}=K_q\,d\mathbf{q}, \label{eq:statics}
\end{equation}
where $J_x$ and $J_q$ are closure Jacobians and $K_q$ is joint-space stiffness. Eliminating $d\mathbf{q}$ gives the analytical mapping used in our network:
\begin{equation}
    K_x
    =
    J_x^\top
    \left(J_q K_q^{-1} J_q^\top\right)^{-1}
    J_x.
    \label{eq:Kx}
\end{equation}

\subsection{AMINN Architecture}
\label{sec:aminn_arch}

We propose AMINN as a cascade analytical-model-informed architecture that integrates a structured analytical mechanics layer into the learning pipeline. The AMINN diagram is illustrated in Fig.~\ref{fig:problem-statement}~(2). For each queried grasp state and external wrench, Stage 1 predicts whether the grasp remains stable, and Stage 2 predicts in-hand tool displacement only for samples that pass the stability prediction, with the analytical mechanics layer embedded in both stages to enforce physically grounded structure. By coupling learned representations with differentiable mechanics, AMINN produces mechanically meaningful predictions, promotes physical consistency, and reduces non-physical hallucinated behavior. This is especially important in safety-critical robotic manipulation, where reliable load-response prediction is essential for both task success and safe human-robot interaction. 

\paragraph{(1) Inputs and stage outputs.}
For each sample $k$, both stages receive the same observable state
\begin{equation}
\mathbf{z}^{(k)}=\left(\mathbf{q}^{(k)},\,\mathbf{x}_{HT}^{(k)},\,\mathbf{w}_e^{(k)},\,\mathcal{C}^{(k)},\,J_x^{(k)},\,J_q^{(k)}\right).
\end{equation}
Here, $\mathbf{q}$ is the hand joint configuration, $\mathbf{x}_{HT}$ is the nominal hand-to-tool pose, $\mathbf{w}_e$ is the applied external wrench, $\mathcal{C}$ is the contact set, and $J_x,J_q$ are the Jacobian terms used by the analytical layer. Stage 1 outputs stability probability $p_{\mathrm{stab}}^{(k)}\in[0,1]$, and Stage 2 predicts in-hand tool displacement $\Delta\hat{\mathbf{x}}_{HT}^{(k)}\in\mathbb{R}^6$ for samples that pass the Stage-1 threshold $\tau$.

\paragraph{(2) Stage 1}
Stage 1 first builds a compact contact summary and fuses it with global state variables. Let $\mathbf{f}$ denote this fused state feature. Let $\Delta\hat{\mathbf{x}}_{\mathrm{aux}}$ and $K_{x,\mathrm{aux}}$ denote the auxiliary displacement and auxiliary Cartesian stiffness from the analytical pathway, and let $\mathbf{g}$ denote tendon-engagement activations. With $h_s$ as the Stage-1 prediction head, the stability output is
\begin{equation}
\hat{s}=\sigma\!\left(h_s\!\left[\mathbf{f},\ \Delta\hat{\mathbf{x}}_{\mathrm{aux}},\ \mathrm{diag}(K_{x,\mathrm{aux}}),\ \mathbf{g}\right]\right),
\end{equation}
where $\mathrm{diag}(\cdot)$ extracts stiffness diagonal terms used as compact mechanics features.

\paragraph{(3) Stage 2}
Stage 2 uses the same mechanics-informed backbone as Stage 1, but it is trained only with ground-truth stable samples. This choice keeps displacement supervision focused on physically meaningful in-hand deformation behavior, instead of mixing in failure-mode samples where displacement labels are less informative for compliant-response learning. During inference, Stage 2 runs only on samples predicted as stable by Stage 1. 

\paragraph{(4) Mechanics head and two-pass analytical layer}
Each stage has its own mechanics head (same architecture, independently trained parameters) that predicts branch-level spring blocks $K_{s,b}$, tendon-direction vectors $\mathbf{d}_b$, tendon gains $k_b$, and gate parameters $(\alpha_b,\beta_b,\rho_b)$ for each branch index $b\in\mathcal{B}$. Spring blocks are constrained SPD and assembled as
\begin{equation}\label{eqn:K_spring}
K_{\mathrm{spring}}=\mathrm{blkdiag}\!\left(\{K_{s,b}\}_{b\in\mathcal{B}}\right),\qquad K_{s,b}\succ0.
\end{equation}
This defines the passive spring contribution before tendon engagement. Let $I$ denote the identity matrix, and let $\epsilon_M,\epsilon_{K_x},\epsilon_{J_q}>0$ denote small regularization terms for numerical stability. The analytical layer then uses two passes so tendon activation is inferred from physically meaningful baseline deformation instead of being injected directly.

Pass 1 estimates a spring-only response:
\begin{equation}\label{eqn:K_x_0}
K_x^{(0)} =
J_x^\top\left(J_q K_{\mathrm{spring}}^{-1}J_q^\top+\epsilon_M I\right)^{-1}J_x+\epsilon_{K_x}I
\end{equation}
\begin{equation}\label{eqn:delta_x_0}
\Delta \mathbf{x}^{(0)} = \left(K_x^{(0)}\right)^{-1}\mathbf{w}_e,\quad
d\mathbf{q}^{(0)}=-\left(J_q+\epsilon_{J_q}I\right)^{-1}J_x\Delta \mathbf{x}^{(0)}.
\end{equation}
Equation~\ref{eqn:K_x_0} maps joint-space stiffness to baseline Cartesian stiffness, and Eqn.~\ref{eqn:delta_x_0} computes baseline in-hand displacement under wrench $\mathbf{w}_e$ and the corresponding joint deflection.

For each branch $b$, tendon gating and tendon stiffness are updated through the following three equations:
\begin{align}
\eta_b &= \langle d\mathbf{q}_b^{(0)},\mathbf{d}_b\rangle,\\
g_b &= \sigma\!\left(\alpha_b(\eta_b-\beta_b)+\rho_b\right),\\
K_{\mathrm{tendon},b} &= (k_b g_b)\,\mathbf{d}_b\mathbf{d}_b^\top.
\end{align}
Here, $\eta_b$ is tendon-direction alignment from pass-1 joint deflection, and $g_b\in[0,1]$ is the tendon-engagement level for branch $b$ (near 0: mostly inactive; near 1: strongly engaged). The branch tendon terms and the final effective spring terms are assembled:
\begin{equation}
K_{\mathrm{tendon}}=\mathrm{blkdiag}\!\left(\{K_{\mathrm{tendon},b}\}_{b\in\mathcal{B}}\right)
\end{equation}
\begin{equation}
K_q^{\mathrm{eff}}=K_{\mathrm{spring}}+K_{\mathrm{tendon}}.
\end{equation}
Pass 2 then recomputes final stiffness and displacement used by the cascade outputs:
\begin{equation}
K_x=
J_x^\top\left(J_q \left(K_q^{\mathrm{eff}}\right)^{-1}J_q^\top+\epsilon_M I\right)^{-1}J_x+\epsilon_{K_x}I
\end{equation}
\begin{equation}
\Delta\hat{\mathbf{x}}_{HT}=K_x^{-1}\mathbf{w}_e.
\end{equation}

\subsection{Loss Function}
\label{sec:aminn_loss}
We train AMINN in two stages to match the cascade architecture: a stability stage and a displacement stage. Stage 1 is trained on all samples for binary stable/unstable prediction, while Stage 2 is trained only on stable samples so displacement learning focuses on physically meaningful in-hand deformation cases.

\paragraph{(1) Stage-1 stability objective}
The Stage-1 classifier is trained with binary cross-entropy on logits:
\begin{equation}
\mathcal{L}_{\mathrm{stab}}
=
\mathrm{BCEWithLogits}(\hat{s}_{\mathrm{logit}}, s).
\end{equation}
This objective directly supervises stability prediction and defines the front-end filter of the cascade. Although this stage does not add an explicit physics penalty term, the logits are computed from mechanics-informed features produced by the analytical layer (auxiliary displacement, stiffness-diagonal terms, and tendon-gate activations), so the Stage-1 gradients still shape the analytical pathway.

\paragraph{(2) Stage-2 displacement objective (stable-only)}
On stable samples, Stage 2 minimizes a unified objective that combines supervised pose fitting 
and structural regularization:
\begin{equation}
\mathcal{L}_{\mathrm{disp}}
=
\mathcal{L}_{\mathrm{pose}}
\,+\,
\mathcal{L}_{\mathrm{str}}.
\end{equation}
The supervised pose-fit terms are $\mathcal{L}_{\mathrm{pose}} = \lambda_{\mathrm{tr}}\mathcal{L}_{\mathrm{tr}} + \lambda_{\mathrm{rot}}\mathcal{L}_{\mathrm{rot}}$
where translation is penalized in Euclidean space and rotation is penalized using geodesic angular error.

The structural regularizers are
\begin{equation}
\mathcal{L}_{\mathrm{str}} = \lambda_{\mathrm{gate}}\mathcal{L}_{\mathrm{gate}} + \lambda_{\mathrm{ten}}\mathcal{L}_{\mathrm{ten}}
\end{equation}
\begin{equation}
\mathcal{L}_{\mathrm{gate}}=\mathrm{mean}\!\left(\mathbf{g}\odot(1-\mathbf{g})\right)
\end{equation}
\begin{equation}
\mathcal{L}_{\mathrm{ten}}=\mathrm{mean}\!\left(K_{\mathrm{tendon}}\odot K_{\mathrm{tendon}}\right).
\end{equation}
Here, $\mathbf{g}$ denotes the tendon-gate activation from the analytical layer (a sigmoid output inferred from baseline tendon-direction stretch in the first pass), with values in $[0,1]$ indicating how strongly each tendon branch is engaged.
The gate term promotes confident on/off activation, while the tendon-energy term limits excessive tendon stiffness magnitude and helps keep learned stiffness adaptation well conditioned.

\paragraph{(3) Cascade operating-point selection}
After model training, inference settings are selected on validation data to balance stability-filter quality and downstream displacement accuracy. In practice, we choose an operating point that preserves strong stability performance while minimizing displacement error on predicted-stable samples.

\begin{figure}[!t]
	\centering
	\includegraphics[width=0.75\columnwidth]{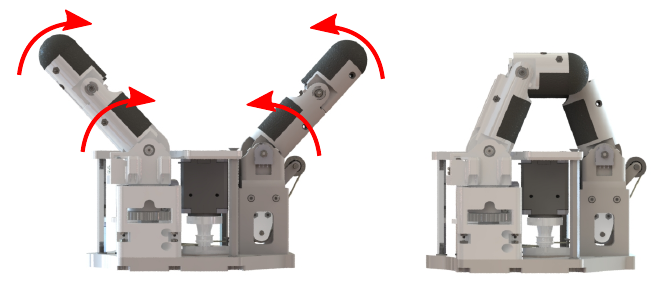}

	\rule{0.85\columnwidth}{0.45pt}
	\vspace{0.5em}
	\hspace{10mm}
	\includegraphics[width=0.75\columnwidth]{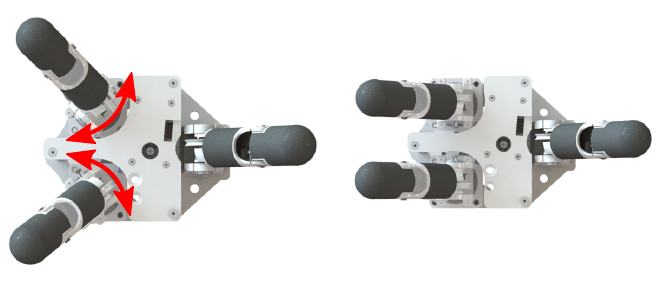}
	\caption{Custom three-finger hand joint motions. Top: flexion at finger joints. Bottom: abduction/adduction at base joints.}
	\label{fig:hand-joint-intro}
\end{figure}

\begin{figure}[b]
	\centering
	\includegraphics[width=0.55\columnwidth]{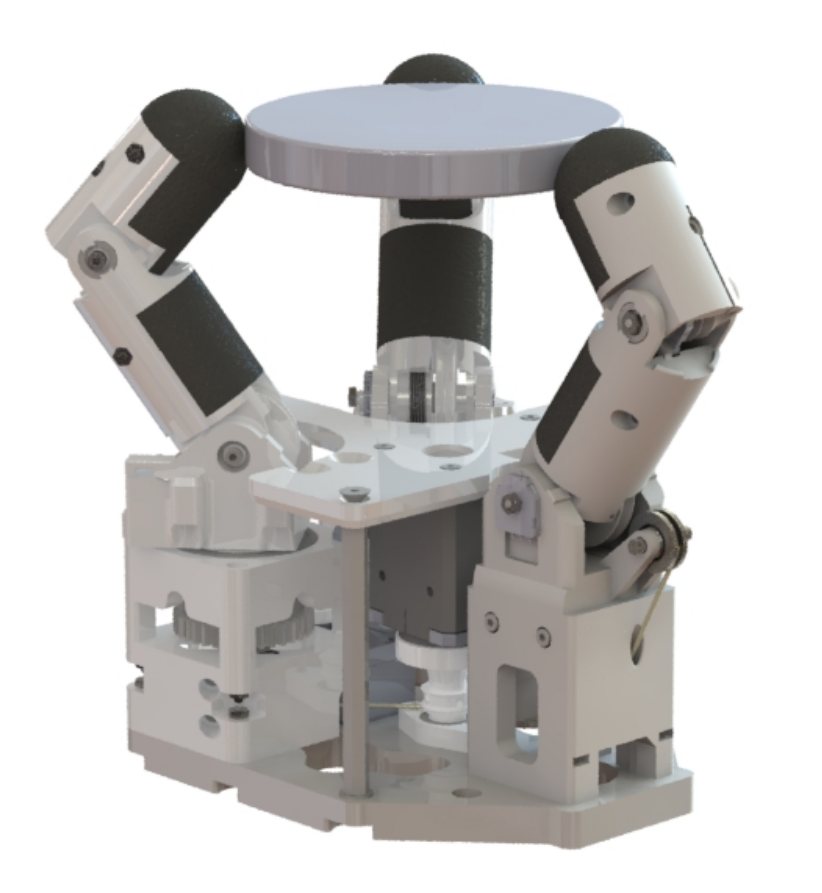}
	\caption{A robotic hand grasping a tool represented as a parallel mechanism.}
	\label{fig:hand-3d}
\end{figure}

\section{Experimental Platform: Underactuated Robotic Hand} \label{sec:nasa_hand}

This section introduces the underactuated robotic hand used to evaluate AMINN. The platform is a tendon-driven, dexterous hand with strong configuration-dependent compliance, making it a representative system for wrench-to-displacement prediction in forceful tool-use settings.

\subsection{Mechanical Design}

The experimental platform (Fig.~\ref{fig:hand-joint-intro}) is a custom three-finger underactuated hand designed and manufactured in-house. The mechanism has eight actuated/measured joint coordinates in total: two mirrored fingers with three coordinates each and one thumb with two coordinates. Each finger uses tendon-driven flexion with spring return. Mechanically, tendon routing provides unilateral load-dependent resistance, while torsional springs provide passive restoring behavior. This coupling is the key source of the two-regime compliance behavior modeled in Sec.~\ref{sec:aminn_arch}.

\subsection{Kinematic Analysis and Parallel-Jacobian Representation}
\label{sec:nasa_hand_kinematics}

\begin{figure}[!b]
	\centering
	\includegraphics[clip,width=0.67\columnwidth]{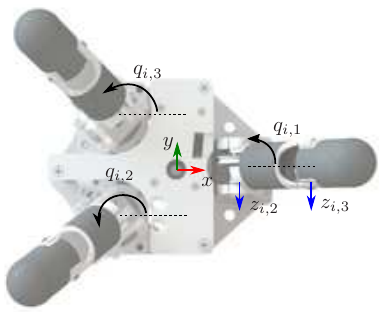}
	\includegraphics[clip,width=0.67\columnwidth]{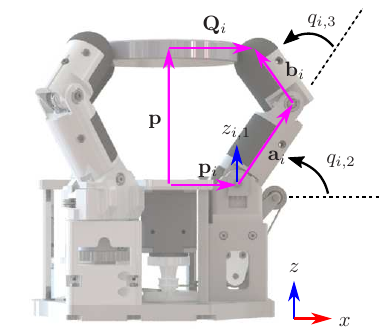}
	\caption{Notation of the three-finger hand used in kinematic analysis.}
	\label{fig:hand-kinematics}
\end{figure}

During grasping, the tool and fingers form a closed-chain system that can be modeled as a parallel mechanism, where the tool acts as the moving platform and the palm as the base (Fig.~\ref{fig:hand-3d}). Let $\dot{\mathbf{x}}=[\dot{\mathbf{p}}^\top,\ {}^{w}\boldsymbol{\omega}_p^\top]^\top\in\mathbb{R}^6$ denote tool twist and let $\dot{\mathbf{q}}$ denote hand joint-rate coordinates.

For branch $i$, the loop-closure relation is
\begin{equation}
\mathbf{p} + {}^{w}R_{p}\mathbf{Q}_i = \mathbf{p}_i + \mathbf{a}_i + \mathbf{b}_i,
\label{eq:loop-closure}
\end{equation}
where $\mathbf{p}$ and ${}^{w}R_p$ are tool position and orientation, $\mathbf{Q}_i$ is branch attachment on the tool, and $\mathbf{p}_i,\mathbf{a}_i,\mathbf{b}_i$ are branch geometry terms.

Differentiating Eq.~\eqref{eq:loop-closure} gives
\begin{equation}
\dot{\mathbf{p}} + {}^{w}\boldsymbol{\omega}_p \times {}^{w}R_{p}\mathbf{Q}_i
=
{}^{w}\boldsymbol{\omega}_{a_i}\times\mathbf{a}_i
+
{}^{w}\boldsymbol{\omega}_{b_i}\times\mathbf{b}_i.
\label{eq:diff-loop-closure}
\end{equation}
Using branch joint-rate coordinates $\dot{\mathbf{q}}_i$, this can be written in linear form
\begin{equation}
\mathbf{A}_i\dot{\mathbf{x}}=\mathbf{B}_i\dot{\mathbf{q}}_i,
\end{equation}
with
\begin{equation}
\mathbf{A}_i =
\begin{bmatrix}
1 & 0 & 0 & 0 & ({}^{w}R_{p}\mathbf{Q}_i)_{z} & -({}^{w}R_{p}\mathbf{Q}_i)_{y} \\
0 & 1 & 0 & -({}^{w}R_{p}\mathbf{Q}_i)_{z} & 0 & ({}^{w}R_{p}\mathbf{Q}_i)_{x} \\
0 & 0 & 1 & ({}^{w}R_{p}\mathbf{Q}_i)_{y} & -({}^{w}R_{p}\mathbf{Q}_i)_{x} & 0
\end{bmatrix},
\end{equation}
\begin{equation}
\mathbf{B}_i =
\begin{bmatrix}
\hat{z}_{i,1}\times(\mathbf{a}_i+\mathbf{b}_i) &
\hat{z}_{i,2}\times(\mathbf{a}_i+\mathbf{b}_i) &
\hat{z}_{i,3}\times\mathbf{b}_i
\end{bmatrix}.
\end{equation}

Stacking all branch equations yields the global closure form
\begin{equation}
J_x\dot{\mathbf{x}}+J_q\dot{\mathbf{q}}=0,
\end{equation}
which is the velocity-level form used in Sec.~\ref{sec:nn} for stiffness mapping. In implementation, branch coordinates are arranged consistently across fingers, and missing coordinates (e.g., the thumb base rolling DoF) are handled by masking in $J_q$.

\subsection{Sensorization}

\begin{figure}[t]
	\centering
		\includegraphics[width=0.75\columnwidth]{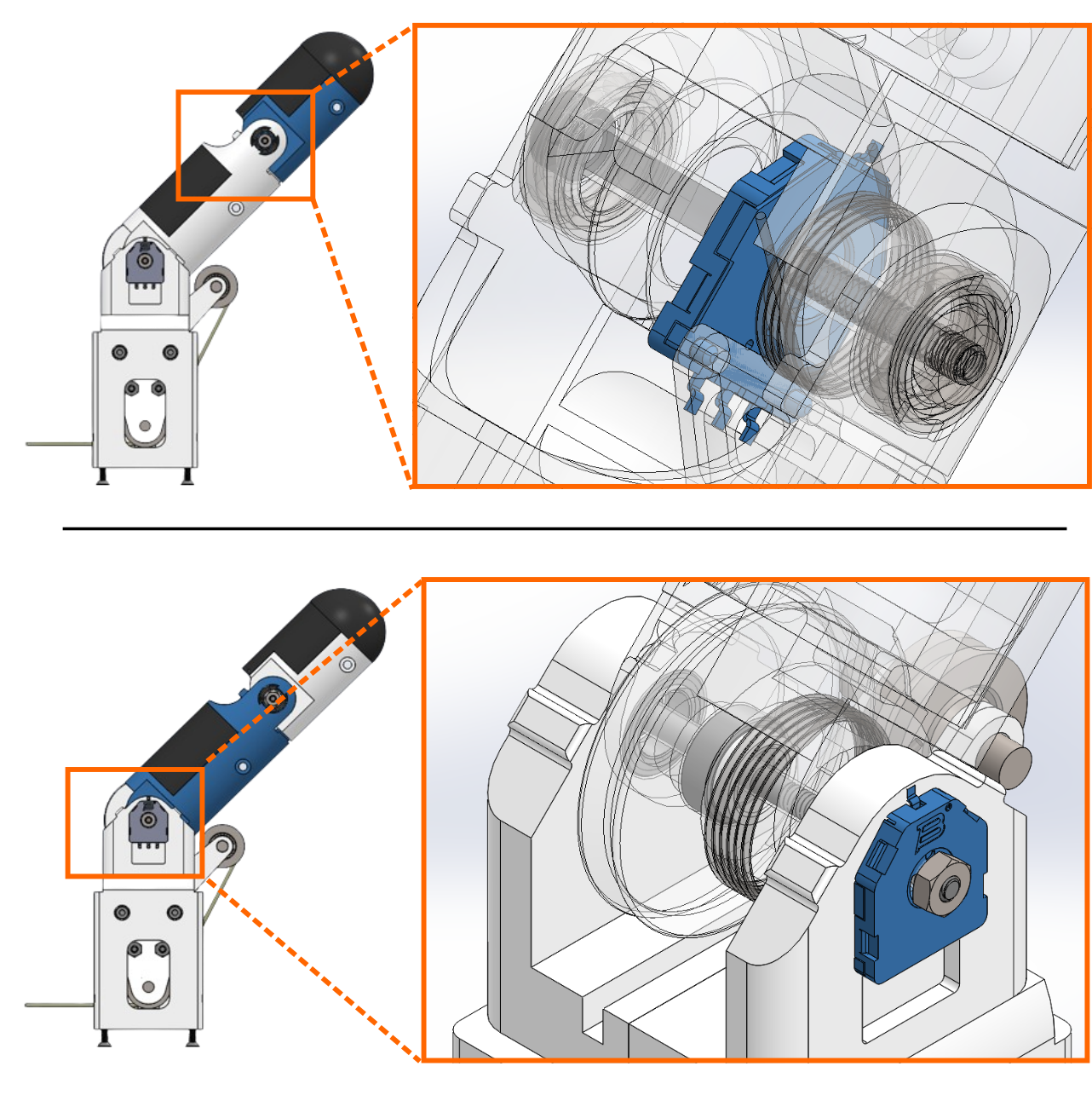}
	\caption{Finger joint sensors and transmission details of the underactuated hand.}
	\label{fig:robotic-hand-finger-details}
\end{figure}

The hand is instrumented to observe the quantities required by the analytical-model-informed pipeline. Joint sensors (Fig.~\ref{fig:robotic-hand-finger-details}) provide $\mathbf{q}$, while a wrist-mounted six-axis force/torque sensor provides $\mathbf{w}_e$ during grasp loading.

Together with contact and kinematic descriptors, these measurements provide the model inputs defined in Sec.~\ref{sec:aminn_arch}. This sensing setup enables synchronized supervision of both stability and in-hand displacement under forceful interaction.

\section{Simulation-Based Data Synthesis for Compliant Tool Grasping}

Building a large-scale dataset for compliant tool grasping is challenging for two reasons. First, direct hardware collection is expensive and cannot practically cover the range of grasp states and load conditions required for robust learning. Second, naive random sampling in high-dimensional joint space produces many infeasible or task-irrelevant poses that violate kinematic, actuation, or contact constraints.

To address these issues, we use a physics-enabled simulation pipeline that generates feasible, task-aligned grasp states and corresponding load responses. The key idea is to constrain sampling through a low-dimensional synergy subspace while preserving physically grounded wrench--displacement behavior through simulation.

\subsection{Physics-Based Simulation Environment}
We implement a simulator consistent with the tendon-driven underactuated architecture and the parallel-robot kinematic model in Sec.~\ref{sec:nasa_hand_kinematics}. The simulated hand has 8-DoF, driven by two position actuators and six custom tendon actuators, and a 6-DoF base actuator that positions the wrist.

The model includes tendon routing, passive joint compliance (e.g., torsional springs), and contact dynamics so that configuration- and load-dependent stiffness behavior emerges naturally. For each rollout, we record joint states, hand-relative tool pose, Jacobians, contact descriptors, and external wrench, providing the quantities needed by Sec.~\ref{sec:nn}.

\subsection{Hand Synergy--Inspired Subspace} \label{subsec:teleop-subspace}
\begin{figure}[!b]
\centering
\includegraphics[width=0.99\columnwidth]{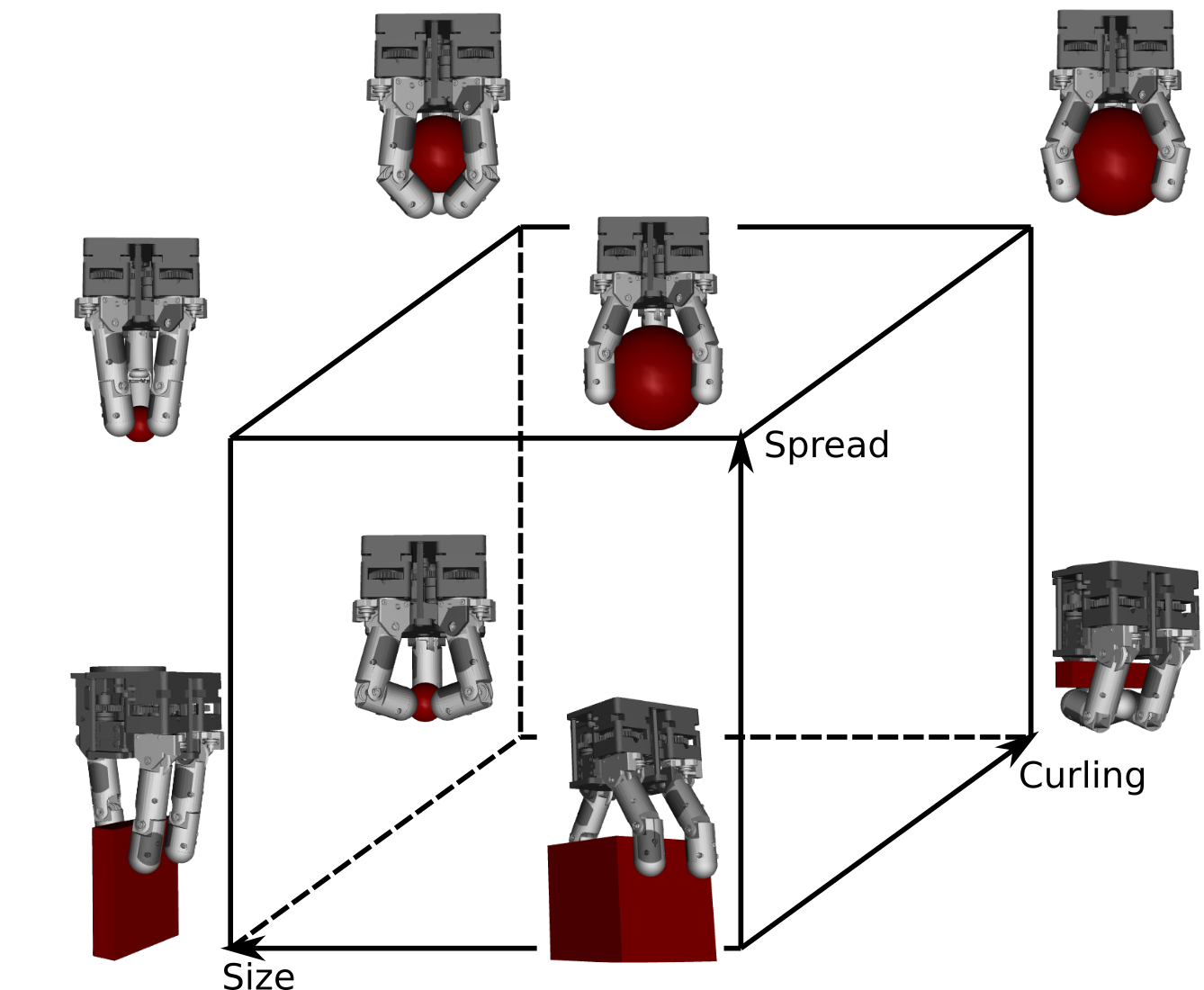}
\caption{Hand synergy-inspired subspace and representative hand grasping configurations.}
\label{fig:teleoperation-subspace}
\end{figure}

Following \cite{CassieTASE}, we use a hand synergy--inspired subspace $\mathbf{T}\in\mathbb{R}^3$ to organize feasible grasp sampling. The three axes represent coordinated motion patterns: size $\sigma$ (open/close), spread $\alpha$ (finger spread), and curl $\epsilon$ (finger curl). This reduces sampling complexity while preserving meaningful grasp variation. An illustration of such a subspace is shown in Fig.~\ref{fig:teleoperation-subspace}.

A hand-specific projection matrix $A\in\mathbb{R}^{N\times 3}$ is defined by synergy basis vectors:
\[
A=\begin{bmatrix}\boldsymbol{\alpha}_H & \boldsymbol{\sigma}_H & \boldsymbol{\epsilon}_H\end{bmatrix},\qquad
\boldsymbol{\alpha}_H,\boldsymbol{\sigma}_H,\boldsymbol{\epsilon}_H\in\mathbb{R}^N.
\]
With neutral pose $\mathbf{o}=\begin{bmatrix}o_1 & \dots & o_N\end{bmatrix}$ and per-axis scaling $\boldsymbol{\delta}\in\mathbb{R}^3$, joint configurations are projected as
\[
\boldsymbol{\psi} = \bigl((\mathbf{q}-\mathbf{o})\cdot A\bigr)\odot\boldsymbol{\delta},\qquad \boldsymbol{\psi}\in\mathbf{T},
\]
where $\odot$ denotes element-wise multiplication.

Intuitively, $A$ encodes how joint coordinates contribute to each synergy mode, $\mathbf{o}$ anchors the mapping at a neutral hand pose, and $\boldsymbol{\delta}$ sets axis ranges. Sampling in $\mathbf{T}$ concentrates on coordinated, physically realizable grasps while avoiding combinatorial sampling in full joint space.

\subsection{Simulation Protocol and Data Acquisition}
For each feasible grasp state from the synergy subspace, we apply randomized disturbance wrenches across direction and magnitude ranges. Each grasp is paired with multiple load cases to cover realistic operating conditions.

We then roll out physics with tendon actuation and contact resolution, and store the model inputs used in Sec.~\ref{sec:nn}:
\[
(\mathbf{q},\ \mathbf{x}_{HT},\ \mathbf{w}_e,\ \mathcal{C},\ J_x,\ J_q).
\]
For supervision, each sample also stores in-hand displacement target $\Delta\mathbf{x}_{HT}$ and a grasp-stability label $s\in\{0,1\}$. The final dataset can be summarized as
\[
\mathcal{D}=\{(\mathbf{q}, \mathbf{x}_{HT}, \mathbf{w}_e, \mathcal{C}, J_x, J_q, \Delta\mathbf{x}_{HT}, s)\},
\]
which aligns with the AMINN architecture in Sec.~\ref{sec:nn} and the evaluation protocol in Sec.~\ref{sec:results}. This dataset is used to compare our proposed hybrid learning method against learning baselines under varied grasping and loading conditions.

\section{Results and Discussion}
\label{sec:results}

\subsection{Experimental Setup and Evaluation Protocol}
We compare our proposed AMINN against a black-box multilayer perceptron baseline (Cascade MLP) under a controlled, fair protocol. Both models receive the same observable inputs, use matched parameter budgets, follow identical training schedules, and are trained and evaluated with the same random seeds.

Both pipelines follow the same two-stage cascade: Stage 1 predicts grasp stability under external loading, and Stage 2 predicts in-hand tool displacement only for samples that pass stability prediction. We report three primary metrics: stability prediction quality via F1, displacement prediction quality via RMSE$_{\mathrm{pose}}$, and a mechanical energy passivity-violation rate. To check physical consistency, we use a simple passivity test on samples predicted as stable. Intuitively, if the external force pushes the tool in one direction, the predicted tool motion should not imply that the system is generating net energy in the opposite direction. We therefore compute a signed-work value from the translational force and predicted translational displacement for each sample. Negative signed work is counted as a passivity violation. The passivity-violation rate is the fraction of predicted-stable samples with violations (lower is better).

\subsection{Quantitative Comparison}
Table~\ref{tab:main_results_v3} summarizes the compact comparison between Cascade AMINN and Cascade MLP at a selected operating point. AMINN achieves strong task-level predictive capability, with competitive stability F1 and displacement RMSE$_{\mathrm{pose}}$ relative to the black-box baseline.

Most importantly, AMINN achieves substantially better energy-based physical consistency in this comparison, with near-zero passivity violations, while the Cascade MLP baseline exhibits a much higher violation rate. This supports the central claim of our hybrid approach: integrating analytical mechanics preserves predictive performance while improving physically consistent behavior.

\begin{table}[t]
\centering
\caption{Comparison between Cascade AMINN and Cascade MLP baseline results.}
\label{tab:main_results_v3}
\begin{tabular}{lcc}
\toprule
 & Cascade AMINN & Cascade MLP \\
\midrule
F1 $\uparrow$ & \textbf{0.96} & 0.95 \\
RMSE$_{\mathrm{pose}}$ $\downarrow$ & \textbf{0.05} & 0.08 \\
Passivity $\downarrow$ & \textbf{0.00} & 0.65 \\
\bottomrule
\end{tabular}
\end{table}

\begin{figure}[t]
\centering
\includegraphics[width=0.9\columnwidth]{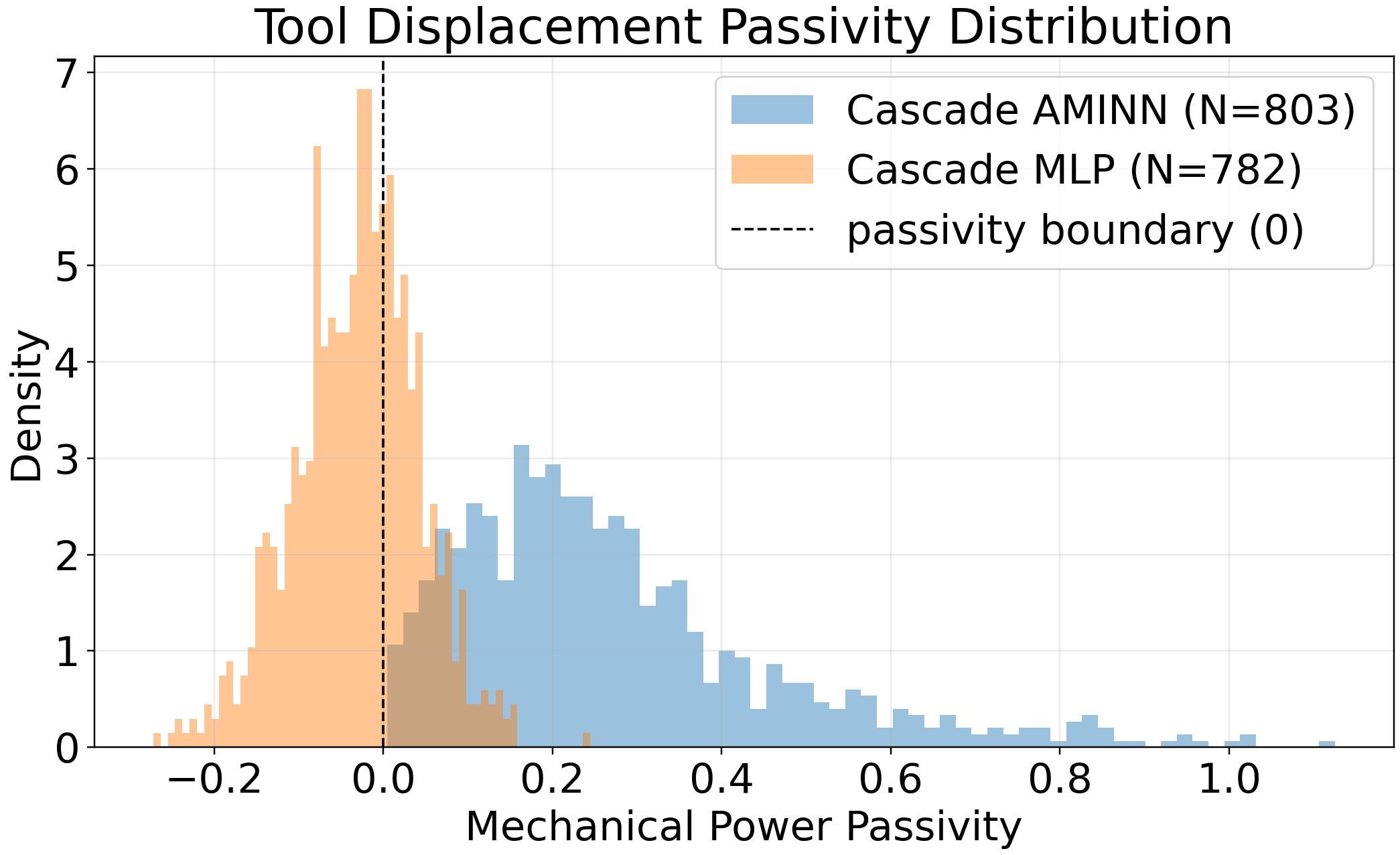}
\caption{Predicted-stable signed-work distribution on the IID test split for Cascade AMINN and Cascade MLP. Signed work is computed per sample as $w=\mathbf{w}_{e}^{\mathrm{tr}}\cdot\Delta\hat{\mathbf{x}}_{HT}^{\mathrm{tr}}$. The dashed line at $w=0$ separates passivity-consistent predictions ($w>0$) from passivity-violating predictions under the translational work proxy ($w<0$).}
\label{fig:passivity_signed_work_v3}
\end{figure}

Figure~\ref{fig:passivity_signed_work_v3} shows the predicted-stable signed-work distribution for both models. The AMINN distribution is concentrated on the positive side with no visible negative-work mass, while the Cascade MLP distribution contains substantial mass below zero. This provides an intuitive view of physics consistency under the translational work proxy.

\subsection{Discussion}
The quantitative results highlight the practical value of hybrid learning for robotic grasping and manipulation. By combining learned mappings with an analytical mechanics layer, AMINN remains task-competitive while grounding its predictions in structured load-response behavior.

A central limitation of purely black-box training is non-interpretability: when predictions fail or drift under changing contact conditions, it is difficult to explain why, diagnose failure modes, or build operator trust. In safety-critical robotics, this can appear as hallucinated, non-physical outputs that are hard to verify before execution. AMINN directly addresses this issue by constraining prediction through mechanics-informed structure, so outputs are physically plausible and physics-consistent under external loading. This improves trustworthiness for deployment, supports safer human-robot operation, and increases the chance of successful grasping and manipulation.

\section{Conclusion}
This paper reframes grasping for underactuated robotic hands from a binary success/failure viewpoint to a compliance-aware prediction problem for forceful tool use. In contact-rich manipulation, grasp retention alone is not sufficient: practical performance also depends on how the grasped tool deforms under external loading, and whether that predicted response is physically plausible.

To address this need, we proposed an analytical model-informed neural network (AMINN), a hybrid learning framework that integrates a structured analytical mechanics layer into the prediction pipeline. AMINN targets two coupled objectives in one model: predicting grasp stability and predicting in-hand tool displacement under load. By embedding mechanics into learning, the model preserves mechanical meaning while adapting to the nonlinear, configuration-dependent behavior of underactuated grasping.

Simulation-based experimental results on a three-finger underactuated robotic hand show that AMINN maintains strong predictive capability relative to a black-box multilayer perceptron baseline, while achieving substantially better energy-based physical consistency (passivity-related behavior). These results support the central claim of this work: analytical-model-informed hybrid learning can improve trustworthiness without sacrificing practical predictive performance.

More broadly, this work supports a shift from black-box prediction toward physically grounded learned models for robotic manipulation. Such models are better aligned with safety-critical deployment because they provide outputs that are both useful and physically interpretable. A key future direction is to further develop \emph{actionable} interpretability, so model structure can be used not only for explanation but also for practical deployment-time adjustment and reliability management under changing interaction conditions. Extending this perspective to broader interaction regimes, richer contact transitions, and tighter integration with downstream planning and control remains an important next step.

\section*{Funding Data}
This material is based upon work supported by the National Science Foundation (NSF), Civil, Mechanical and Manufacturing Innovation (CMMI) Division, under Award ID\# CMMI-2138896.

\end{document}